\documentclass[letterpaper, 10 pt, conference]{ieeeconf} 
\IEEEoverridecommandlockouts
\usepackage{cite}
\usepackage[font=footnotesize]{subfig}
\usepackage{multirow}
\usepackage{ifpdf}
\usepackage[pdftex]{graphicx}
\usepackage{epstopdf}
\usepackage[cmex10]{amsmath}
\usepackage{algorithmic}
\usepackage{array}
\usepackage{mdwmath}
\usepackage{mdwtab}
\usepackage{eqparbox}
\usepackage{verbatim}
\usepackage{amsmath,siunitx}

\usepackage[font=footnotesize]{subfig}
\usepackage{fixltx2e}
\usepackage{stfloats}

\usepackage{amsmath,bm}
\usepackage{amssymb}
\usepackage{color}
\usepackage{subfig}

\usepackage{multirow}
\usepackage{float}
\usepackage{caption}
\usepackage{comment}

\usepackage{makecell}
\usepackage{amsmath}
\usepackage{graphicx}
\usepackage[colorlinks=true, allcolors=blue]{hyperref}
\usepackage{adjustbox}
\usepackage{blindtext}
\usepackage{hyperref}

\usepackage[font=footnotesize]{subfig}
\usepackage{fixltx2e}
\usepackage{stfloats}

\usepackage{amsmath,bm}
\usepackage{amssymb}
\usepackage{color}
\usepackage{subfig}

\usepackage{multirow}
\usepackage{float}
\usepackage{caption}
\usepackage[normalem]{ulem}
\usepackage{comment}
\usepackage{hyperref}
\usepackage{cite}
\usepackage[font=footnotesize]{subfig}

\usepackage{ifpdf}

\usepackage{epstopdf}
\usepackage[cmex10]{amsmath}

\usepackage{array}
\usepackage{mdwmath}
\usepackage{mdwtab}
\usepackage{eqparbox}
\usepackage{verbatim}
\usepackage{booktabs}
\usepackage{multirow}
\usepackage{graphicx}
\usepackage{algorithm}
\usepackage{algorithmic}
\usepackage{array} 
\usepackage{algorithm}
\usepackage{algorithmic}
\usepackage{xcolor}

\begin{document}
	\title{\LARGE \bf
Physics-Informed Machine Learning for Efficient Sim-to-Real Data Augmentation in Micro-Object Pose Estimation}

\author{Zongcai Tan*, Lan Wei* and Dandan Zhang
\thanks{*Equal Contribution.}
\thanks{Zongcai Tan, Lan Wei and Dandan Zhang are with the Department of Bioengineering, Imperial-X Initiative, Imperial College London, London, United Kingdom.  Corresponding: d.zhang17@imperial.ac.uk.}
}

\maketitle

\pagestyle{empty}  
\thispagestyle{empty} 

\maketitle
\pagestyle{empty}  
\thispagestyle{empty} 

\begin{abstract}
Precise pose estimation of optical microrobots is essential for enabling high-precision object tracking and autonomous biological studies. 
However, current methods rely heavily on large, high-quality microscope image datasets, which are difficult and costly to acquire due to the complexity of microrobot fabrication and the labour-intensive labelling. 
Digital twin systems offer a promising path for sim-to-real data augmentation, yet existing techniques struggle to replicate complex optical microscopy phenomena, such as diffraction artifacts and depth-dependent imaging.
This work proposes a novel physics-informed deep generative learning framework that, for the first time, integrates wave optics-based physical rendering and depth alignment into a generative adversarial network (GAN), to synthesise high-fidelity microscope images for microrobot pose estimation efficiently.
Our method improves the structural similarity index (SSIM) by 35.6\% compared to purely AI-driven methods, while maintaining real-time rendering speeds (0.022 s/frame). 
The pose estimator (CNN backbone) trained on our synthetic data achieves 93.9\%/91.9\% (pitch/roll) accuracy, just 5.0\%/5.4\% (pitch/roll) below that of an estimator trained exclusively on real data. 
Furthermore, our framework generalises to unseen poses, enabling data augmentation and robust pose estimation for novel microrobot configurations without additional training data.
\end{abstract}












\section{Introduction}
Optical microscopy (OM) is commonly integrated with microrobotic systems for observing and characterizing micro/nano objects, providing essential visual feedback for precise three-dimensional (3D) localization and pose determination~\cite{dong2024ai}. Accurate visual perception of optical microrobots is crucial for biomedical tasks at micro and nanoscales, such as targeted delivery, micromanipulation, and microassembly~\cite{grier2003revolution,zhang2022fabrication,zhang2022micro,sha2019review}. However, microscopic image degradation due to optical defocusing, diffraction, and background noise significantly hinders robust microrobot tracking and pose estimation~\cite{muinos2021reinforcement}. Thus, effective vision-based methods for tracking and pose estimation are critical for enhancing microrobot perception and reliability in biomedical applications.

Due to high costs of micro/nano-fabrication and difficulties in precise out-of-plane pose control, obtaining large, diverse, and well-labelled real-world datasets is technically challenging and expensive~\cite{li2024control}. The scarcity of high-quality annotated data severely restricts the performance of data-driven pose estimation methods~\cite{yang2024machine,shurrab2022self,plompen2020joint}. Zhang~\textit{et al.} proposed a Generative Adversarial Network (GAN)-based augmentation to generate synthetic microrobot images to supplement real-world data~\cite{zhang2022micro}. However, their approach relies solely on learned data distributions, neglecting optical microscopy physics, causing discrepancies between synthetic and real images. Consequently, these images often miss critical optical phenomena necessary for high-fidelity microrobot imaging. Additionally, GANs trained on limited datasets may poorly generalize to unseen configurations, resulting in poor performance when the synthetic data does not fully represent the diversity of microrobot poses.

Most existing physical simulation methods focus on fluorescence and electron microscopy, with limited research dedicated to modeling OM images~\cite{19rogers2012super,22wang2011optical,23zhang2023large,24li2024statistical,balakrishnan2023single}. Current simulation methods face a trade-off: either achieving high physical accuracy with substantial computational costs unsuitable for real-time applications, or simplifying physics, sacrificing generality and reliability~\cite{22wang2011optical,23zhang2023large}. This constraint also hampers high-fidelity synthetic data generation needed to augment experimental datasets for training pose estimation models.

\begin{figure}[t!]
\captionsetup{font=footnotesize,labelsep=period}
\centering
\includegraphics[width=1\hsize]{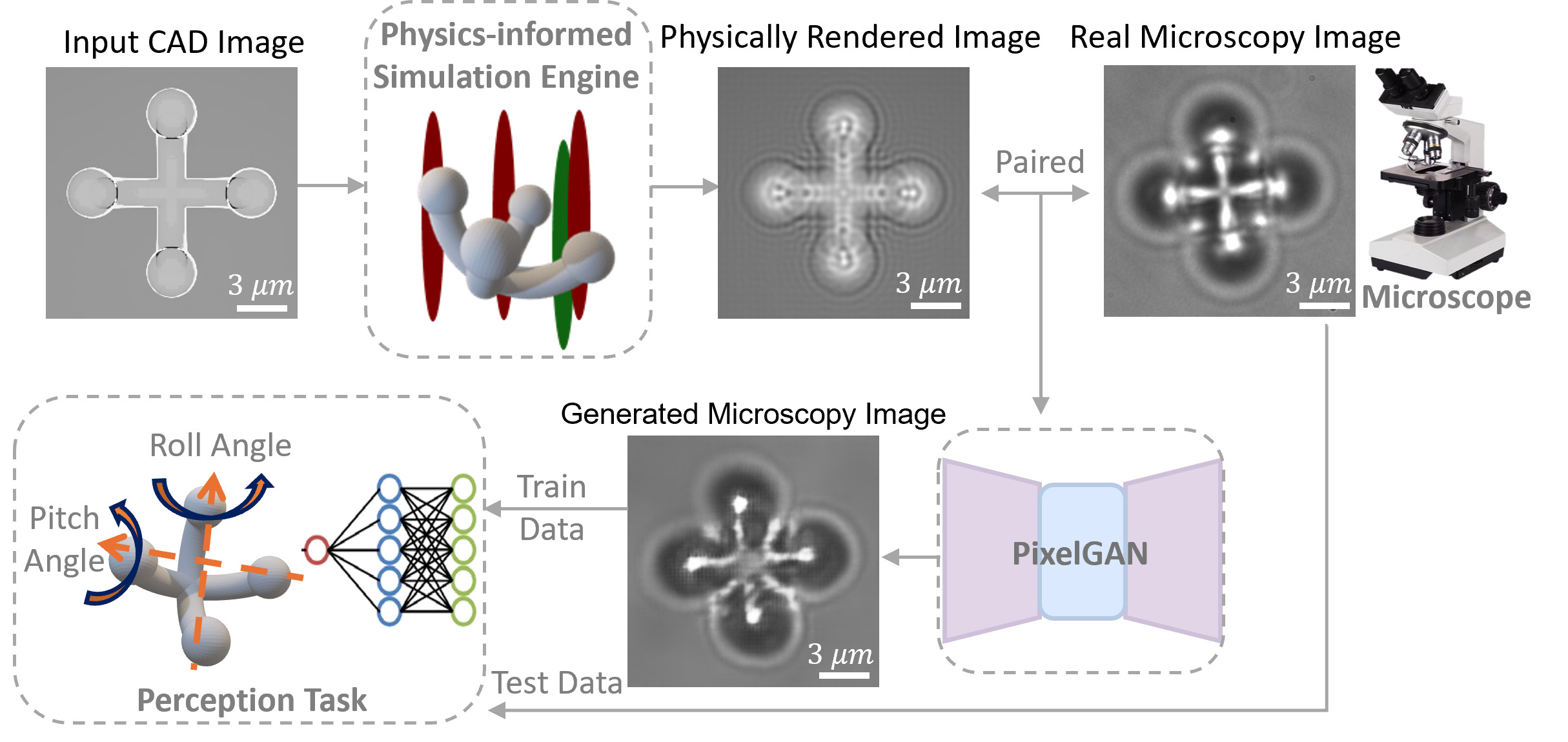}
\caption[Data Alignment]
{Concept overview of the physics-informed machine learning network for efficient sim-to-real microscopy data generation.}
\label{fig-ConceptOverview}
 \vspace{-0.5cm}
\end{figure}%

Here, this work proposes a high-fidelity digital twin system integrating physics-informed machine learning to facilitate sim-to-real OM image generation. Specifically, the system simulates key optical effects such as defocus blur, diffraction rings, and depth-dependent variations by applying wave optics principles, optimizing computational complexity for efficient simulation. Additionally, simulated images are aligned with experimental data using a pixel-to-pixel generative adversarial network (PixelGAN)~\cite{isola2017image}, further reducing discrepancies between datasets. Image fidelity and efficiency are evaluated using the Structural Similarity Index (SSIM), Peak Signal-to-Noise Ratio (PSNR), Mean Squared Error (MSE), and image generation inference time. Models trained on simulated and real datasets were tested on microrobot pose estimation tasks. Results confirm that integrating physics-based modeling with data-driven refinement improves both interpretability and realism, preserving crucial depth-encoding features and enhancing sim-to-real transfer.

In summary, we introduce the first physics-informed PixelGAN framework for developing a digital twin of an OT-actuated, complex-shaped microrobotic system. As illustrated in Fig.~\ref{fig-ConceptOverview}, the primary innovation is integrating physics-based rendering and pixel-level depth alignment into GAN training, significantly enhancing synthetic image realism and fidelity in depth encoding. This method effectively addresses data scarcity for microrobot perception and balances simulation fidelity with computational efficiency.

\vspace{0.2cm}
The \textbf{Main Contributions} of this paper are as follows:
\begin{enumerate}
    \item The work introduces a high-fidelity digital twin system that integrates physics-informed machine learning to simulate OM images of optically actuated microrobots. By incorporating wave optics principles, the model accurately reproduces key optical effects, bridging the gap between simulation and real-world imaging.

    \item To enhance simulation realism, this work employs PixelGAN, a deep generative model that refines simulated images to better align with real experimental data. 
    This approach significantly improves dataset quality for visual perception algorithms, facilitating accurate and robust pose estimation of optical microrobots.

    \item The system achieves real-time generation of high-fidelity microscopic images. By optimizing computational complexity without sacrificing fidelity, it accelerates training efficiency, increases dataset diversity, and reduces data collection and labelling costs, enabling broader microrobot applications in real-time visual tasks.

    \item The proposed framework demonstrates generalisability to unseen pose configurations. PixelGAN-30 (five held-out poses) still yields data that trains high-accuracy estimators on unseen poses (only 2.4\%--2.5\% drop relative to PixelGAN-35), demonstrating robustness.
\end{enumerate}

\section{Related Work}

\subsection{Simulation of OM Imaging for Micro-Objects}

In optical microscopy, noise and artifacts such as diffraction rings and defocus blur often encode critical information about an object's pose and depth. 
Even minor mismatches between simulated and real imaging can introduce significant errors in training visual algorithms. Current research primarily focuses on super-resolution reconstruction \cite{19rogers2012super}, with limited attention to the physical modeling of OM imaging. For example, Nasse and Woehl proposed a point spread function (PSF) model based on vector theory, but it struggles to handle fully 3D imaging of multilayered or complex 3D samples \cite{20nasse2010realistic}. Marian's research on 3D amplitude PSF offers physically based image simulations, but its high computational demand limits real-time applications \cite{21marian2007complex}. Wang et al. employed geometric optics ray tracing, Mie theory, and FDTD to simulate nano-scale microsphere imaging, but these methods face computational challenges, particularly with complex interference and wave effects \cite{22wang2011optical}. Similarly, Zhang et al. introduced a diffractive optical element (DOE)-based approach combined with deep learning for image simulation and reconstruction. While this method simplifies component modeling to reduce complexity, it sacrifices accuracy in representing real optical pathways and assumes a uniform PSF over extended depth ranges, which limits its generalizability in complex samples \cite{23zhang2023large}. Li et al. simulated confocal microscopy imaging focused on fluorescent signals but did not achieve virtual-real synchronization, neglecting continuous complex structures \cite{24li2024statistical}. Overall, significant gaps remain in OM visualization and rendering, particularly in high-fidelity models that support fast and accurate imaging of complex 3D structures.

\subsection{Machine Learning for Micro/Nanorobot Tracking}
Supervised learning methods have been widely explored for micro/nanorobot tracking and 3D perception \cite{cenev2016object}. Deep neural networks have been used to estimate the 3D pose and depth of optically transparent microrobots \cite{grammatikopoulou2019three,zhang2020data} and to localize multiple microrobots for automated manipulation via OT \cite{ren2022machine}. More recently, a large-scale optical microrobot dataset with standardized pose and depth benchmarks has been released, enabling reproducible evaluation and cross-model comparison \cite{wei2025dataset}. Physics-informed frameworks that combine focus-based metrics with convolutional features further improve depth estimation, particularly under limited data conditions \cite{wei2025physics}. Despite these advances, model performance still depends heavily on expensive labeled data, constrained by microrobot fabrication, imaging, and annotation costs. Consequently, data augmentation and data-efficient learning strategies remain key bottlenecks for scaling machine learning–based tracking algorithms.


\section{Methodology}

\begin{figure*}[t!]
  \captionsetup{font=footnotesize,labelsep=period}
  \vspace*{1em} 
  \centering
  \includegraphics[width=1\hsize]{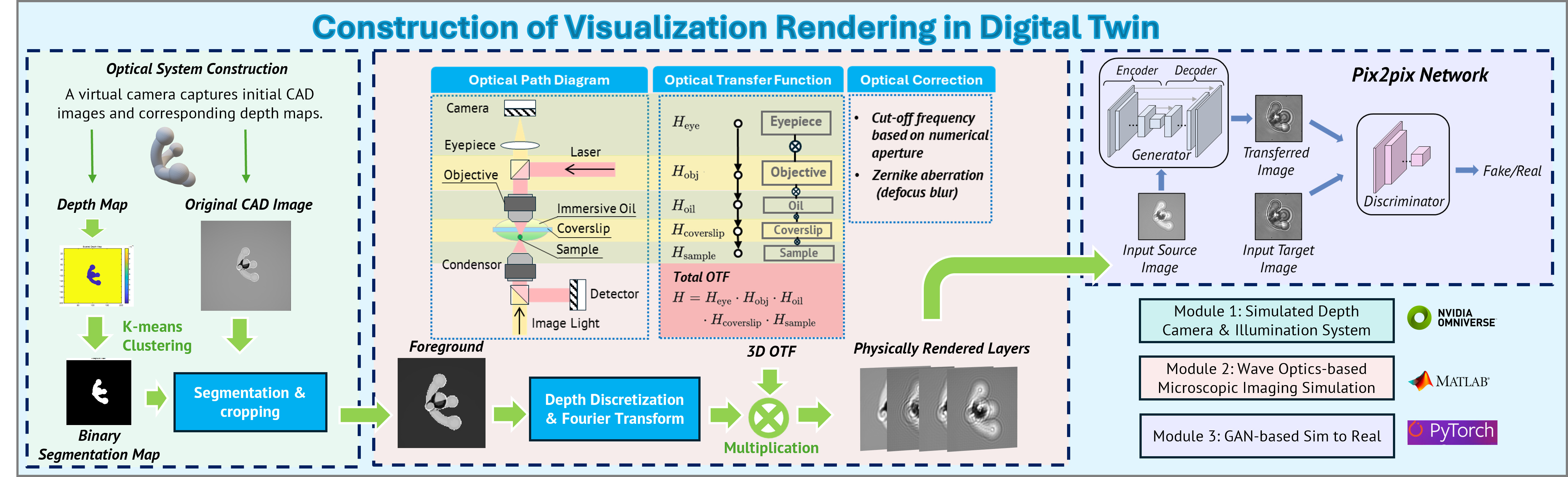}
  \caption[Physical model of the visual rendering]
  {Workflow of the visualization rendering system: A virtual optical microscope system was constructed in Isaac Sim based on real-time optical path parameters and predicted robotic poses. Using the initial CAD images and depth maps captured by a virtual camera, high-fidelity simulated images are generated via the visualization rendering module based on wave optics. The reality gap of the virtual image was further reduced through a sim-to-real module using PixelGAN \cite{isola2017image}.}
  \label{fig-visualModel}
\end{figure*}

\subsection{System Overview and Workflow}
To accurately replicate the imaging process in the OT system, this work develops a high-fidelity microscopy simulation model based on wave optics. This model incorporates the entire optical path of the microscope and accounts for physical factors influencing image quality, such as objective focal length ($f_{\text{obj}} = 50\,\text{mm}$), eyepiece focal length ($f_{\text{eye}} = 20\,\text{mm}$), numerical aperture ($\text{NA} = 1.45$), and illumination wavelength ($\lambda = 632.8\,\text{nm}$). These parameters critically determine the optical resolution, magnification, and overall imaging fidelity. By integrating these components with a deep learning GAN model, the simulation ensures that imaging results within the digital twin environment closely align with real-world experimental data. 

The workflow of the visualization rendering system is shown in Fig.~\ref{fig-visualModel}. It begins with constructing a virtual optical system using the NVIDIA platform, which generates an optical model based on real-time optical path parameters and simulated robotic poses. A virtual camera captures initial CAD images and corresponding depth maps, synchronizing closely with real experimental conditions. The depth map undergoes $k$-means clustering to segment the foreground (microrobot) from the background, allowing for image cropping that fully encloses the robot, significantly reducing computational load.

Once the foreground is segmented, image formation begins by deriving the optical transfer function (OTF) of the microscope using a comprehensive wave-optics model of the system's optical path. Subsequently, the robot depth is discretized along the z-axis into multiple layers using the depth map. Each depth layer is transformed into the Fourier frequency domain, enabling efficient multiplication with its corresponding OTF, instead of traditional spatial-domain convolution. During this frequency-domain processing, a NA cutoff is applied by setting the OTF values to zero for spatial frequencies that exceed the microscope’s maximum resolvable frequency ($f_{\text{cutoff}}$). This effectively removes non-physical frequencies from the simulation, ensuring realistic modeling of optical limits. Parseval's theorem is applied throughout to ensure energy conservation and signal integrity during the convolution process. Finally, to bridge the simulation-to-reality gap, rendered images undergo refinement using a sim-to-real module based on PixelGAN~\cite{isola2017image}, reducing visual discrepancies and ensuring alignment with experimental data.


\subsection{Optical System and Wave Optics Simulation}
To simulate the microscope's optical system accurately, a detailed optical path model was constructed, as depicted in Fig.~\ref{fig-visualModel}. This model encompasses critical optical components: objective lens, eyepiece, cover slip, immersion oil, and sample medium (deionized water). Incorporating precise optical properties such as refractive indices—immersion oil ($n_{\text{oil}}=1.515$), coverslip ($n_{\text{coverslip}}=1.515$), and sample medium ($n_{\text{sample}}=1.33$)—is crucial. Variations in refractive index significantly affect the NA, diffraction patterns, and effective wavelength within different media, influencing the achievable resolution and imaging accuracy.

Fourier optics was employed to simulate light propagation accurately, with each optical component represented by an OTF that encapsulates its impact on the propagating wavefront. Key transfer functions for optical elements are expressed as follows:
\begin{itemize}
    \item \textbf{Eyepiece OTF}:
    \begin{equation}
    H_{\text{eye}} = \exp\left(-i\pi\frac{(U^2 + V^2)\lambda_e}{f_{\text{eye}}}\right)
    \label{eq:otf_eye}
    \end{equation}

    \item \textbf{Objective OTF}:
    \begin{equation}
    H_{\text{obj}} = \exp\left(-i\pi\frac{(U^2 + V^2)\lambda_e}{f_{\text{obj}}}\right)
    \label{eq:otf_obj}
    \end{equation}

    \item \textbf{Cover Slip OTF}:
    \begin{equation}
    H_{\text{coverslip}} = \exp\left(i2\pi\lambda_{\text{coverslip}}(U^2 + V^2)\right)
    \label{eq:otf_coverslip}
    \end{equation}
\end{itemize}

The combined effect of these components yields the total system OTF:
\begin{equation}
H_{\text{total}} = H_{\text{eye}} \times H_{\text{obj}} \times H_{\text{coverslip}} \times H_{\text{oil}} \times H_{\text{sample}}
\label{eq:total_otf}
\end{equation}

The NA cutoff frequency, representing the highest spatial frequency resolvable by the optical system, is calculated as:
\begin{equation}
f_{\text{cutoff}} = \frac{\text{NA}\cdot n_{\text{oil}}}{\lambda}
\label{eq:cutoff}
\end{equation}
Frequencies exceeding this limit are excluded to ensure the simulated image adheres strictly to the physical constraints of the microscope.

Optical aberrations due to lens imperfections were modeled using Zernike polynomials. Specifically, the fourth-order Zernike polynomial ($Z_4$), which describes primary spherical aberration (a common lens aberration causing blurred focal spots), was applied directly as an additional phase shift in the frequency domain:
\begin{equation}
H_{\text{total}}(U,V) \rightarrow H_{\text{total}}(U,V) \times \exp(i Z_4), \quad Z_4 = \sqrt{3}(2\rho^2 - 1)
\label{eq:zernike}
\end{equation}
where $\rho$ is the normalized radial coordinate, defined as the radial distance from the optical axis normalized by the maximum radius set by the NA cutoff.

\begin{figure*}[t!]
  \captionsetup{font=footnotesize,labelsep=period}
\vspace*{1em} 
\centering
\includegraphics[width=0.95\hsize]{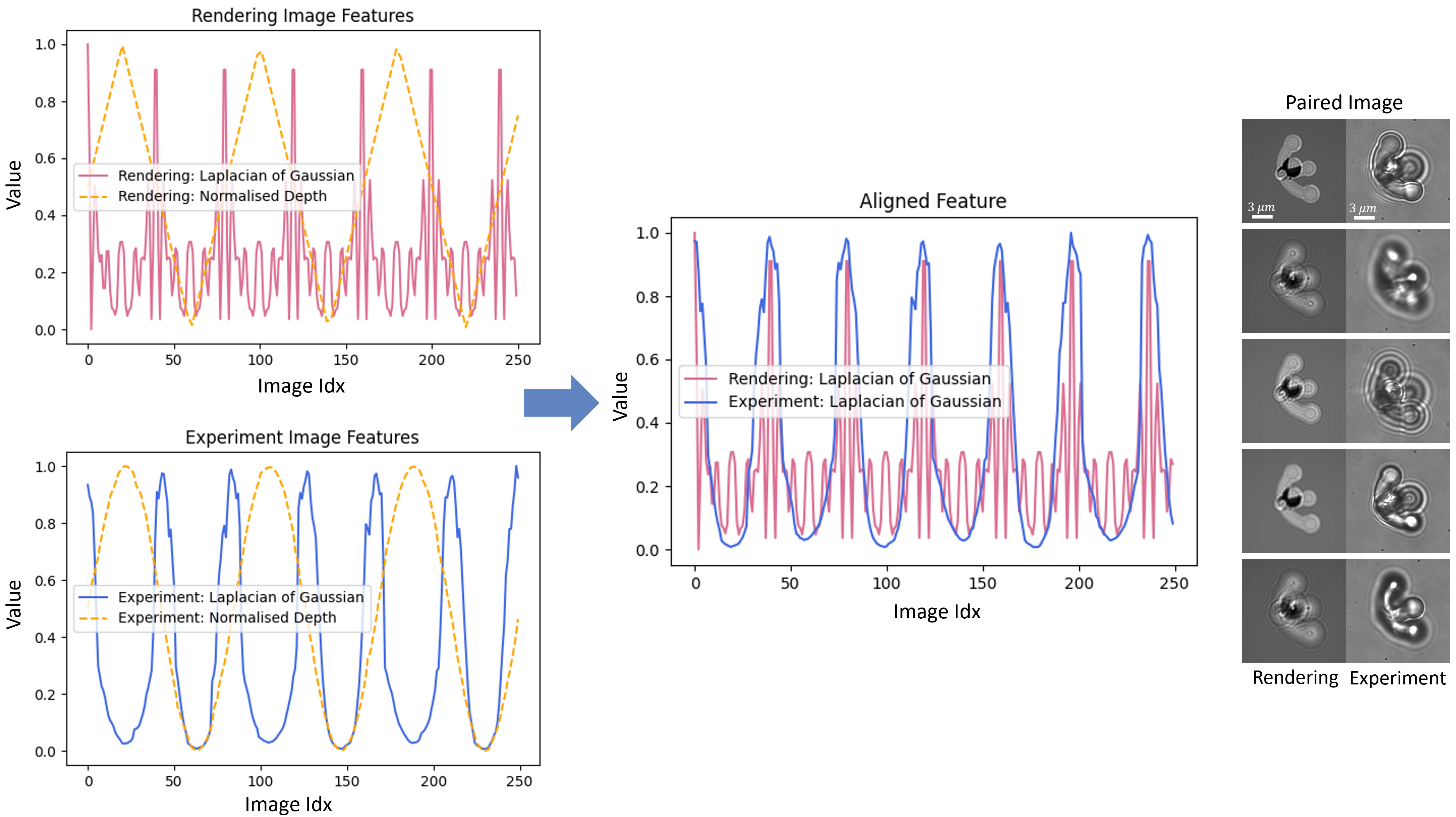}
\caption[Data Alignment]
{Alignment of rendering and experiment image features based on Laplacian of Gaussian (LoG) analysis.
LoG values and normalised depth are extracted for each dataset (left).
Peak LoG frames (corresponding to the focal plane) are used to segment the datasets.
To enable one-to-one pairing, data within each segment is balanced (middle), facilitating aligned image pairs for downstream training (right).
}
\label{fig-aligned_process}
\vspace{-0.5cm}
\end{figure*}%

\subsection{Depth Discretization and 3D Visual Rendering}
As shown in Fig.\ref{fig-visualModel}, to generate high-fidelity microscopic images, clear images and depth maps of the sample were obtained using Isaac Sim and transmitted to MATLAB via ROS. 
The microscope's focal plane was defined as the 0\,\si{\micro\meter} reference. The robot's operational depth, spanning from $-10$\,\si{\micro\meter} to $+10$\,\si{\micro\meter}, was uniformly partitioned into 40 discrete layers, with each layer corresponding to a distinct robotic position and its respective optical projection.
The angular spectrum propagation method was then used to calculate the OTF for each depth layer, and physically rendered images were generated by convolving the OTF with the corresponding depth robot image. Additionally, to improve real-time performance, parallel computing and GPU acceleration were employed, and Parseval's theorem was applied to ensure energy conservation across Fourier transformations.


\subsection{Micro-Fabrication and Experimental Setting}
The microrobots were fabricated using IP-L Photoresist (Nanoscribe, Germany) with a Nanoscribe 3D printer (Nanoscribe GmbH, Germany) through a two-photon polymerisation (2PP) process \cite{kawata2001finer}. 
The microrobots used for collecting experimental data were printed on glass substrates and placed on a deionized water spacer (DI) \cite{4zhang2020distributed}.
The experimental setup consists of an OT (Elliot Scientific, UK) integrated with a nanopositioner (Mad City Labs Inc.). 
The collected dataset comprises microscope images recorded via a CCD camera, capturing the depth of microrobot and their poses.  The pitch angle (`P') refers to the pose of the robot along the $x$-axis, while the roll angle (`R') refers to the pose along the $y$-axis.
As $10^{\circ}$ is set as the resolution of change angle, the study has a total of 35 different kinds of pose classes for the optical microrobot designed by Zhang et al.~\cite{4zhang2020distributed}. 
Each image frame has a resolution of 678×488 pixels. 
To obtain depth values, the printed microrobots were attached to a glass plane and positioned on a piezoelectric substrate, generating the $z$-axis trajectories.


 

\begin{figure*}[t!]
  \captionsetup{font=footnotesize,labelsep=period}
\vspace*{1em} 
\centering
\includegraphics[width=0.85\hsize]{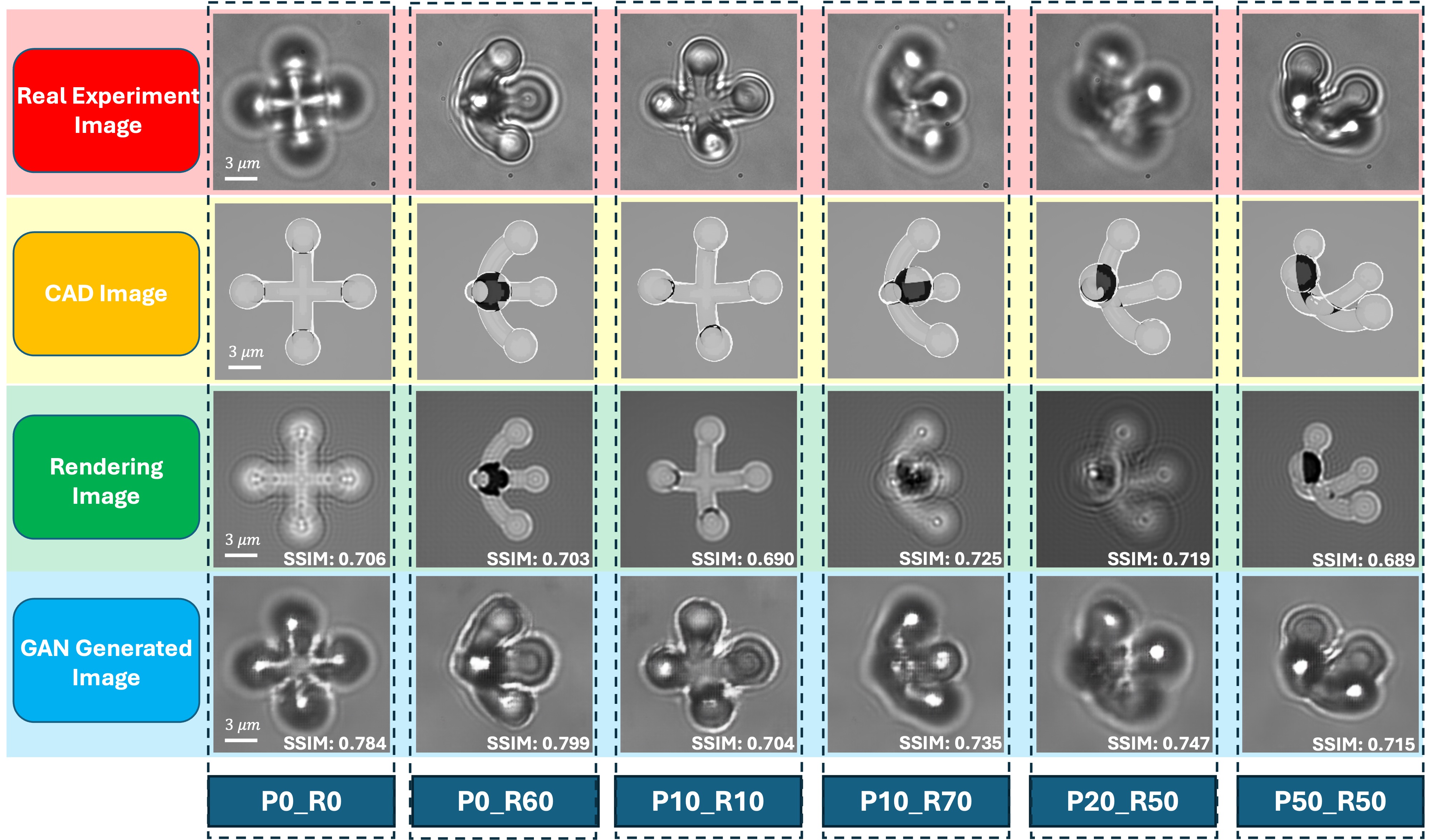}
\caption[Data Alignment]
{Qualitative evaluation of image generation methods across varying poses and depths, demonstrating the visual fidelity of simulated microscope images compared to real experimental images. The comparison includes real experimental images (red), CAD renderings (yellow), physically rendered images (green), and GAN-generated images (blue). 
}
\label{fig-result}
\vspace{-0.2cm}
\end{figure*}%

\subsection{Sim-to-Real Transfer}

Although physics-based visualization rendering models can generate virtual microscope images that preserve depth information, they often exhibit discrepancies from real experimental images—particularly in terms of contrast and gloss. 
To address this, the work employs a PixelGAN further to reduce the gap between simulated and real images. 
This work first aligns features between physically rendered images and their corresponding experimental images at the same depths. 
Because the sharpness of an object’s edges varies with depth under a microscope, edge detection becomes a key factor in this alignment. 
The Laplacian operator, as a second-order derivative, effectively identifies edges by capturing rapid changes in intensity~\cite{wang2007laplacian}, and combining it with Gaussian smoothing provides a robust feature for the alignment process.

First, the Laplacian of Gaussian (LoG) values are computed for each image in both datasets as shown in the left part of Fig.\ref{fig-aligned_process}.
The image corresponding to the peak LoG value is identified as the frame located at the microscope's focal plane, where the normalised depth is 0, and the edges of the target object are the sharpest.
Using the peak LoG points from the physically rendered images and experimental images, the datasets are divided into multiple segments.
For each segment, the number of physically rendered images and experimental images is balanced by randomly selecting data points from the segment with more data. 
This ensures that the physically rendered images and experimental images have the same number of samples, allowing them to be paired one-to-one for the subsequent PixelGAN training.

PixelGAN consists of two main components: a Generator ($G$) and a Discriminator ($D$) \cite{isola2017image}. It learns a mapping from the physics-rendered image $x$ and a random noise vector $z$ to the corresponding real experimental image $y$, formulated as: $\{x, z\} \rightarrow y$.
The generator adopts a U-Net-style architecture, designed to synthesize realistic images that closely resemble real microscopy images. Meanwhile, the discriminator, implemented as a PatchGAN, operates adversarially to distinguish between real and generated images at the patch level, rather than evaluating entire images. This patch-based approach helps preserve fine-grained textures and structural consistency in the generated images.
During training, the generator is optimized to fool the discriminator, while the discriminator simultaneously improves its ability to identify synthetic images, creating a competitive learning process. This training pipeline is illustrated in the right part of Fig.~\ref{fig-visualModel}.
The objective of the PixelGAN can be expressed as:
\begin{equation}
\begin{aligned}
\mathcal{L}(G, D)= & \mathbb{E}_{x, y}[\log D(x, y)]+ \\
& \mathbb{E}_{x, z}[\log (1-D(x, G(x, z))],
\end{aligned}
\end{equation}
where $G$ tries to minimize the objective against an adversarial $D$ that tries to maximize it, i.e. $G^* = \text{arg}\,\text{min}_G\,\text{max}_D\,\mathcal{L}(G,D)$.
The work further adds an L1 distance to help the generator reduce blurring:
\begin{equation}
\mathcal{L}_{L 1}(G)=\mathbb{E}_{x, y, z}\left[\|y-G(x, z)\|_1\right].
\end{equation}
The final objective is:
\begin{equation}
G^*=\arg \min _G \max _D \mathcal{L}(G, D)+\lambda \mathcal{L}_{L 1}(G) .
\end{equation}
In the implementation, the work sets $\lambda$ as the default value 100.

\section{Experiments and Results}
\subsection{Data}
The aligned data used for PixelGAN training consists of 15,820 images (each consisting of one physically rendered image and one experiment image), corresponding to 35 sets of optical microrobots with different poses~\cite{wei2025dataset}. 
The resulting paired data are shown in the right part of Fig. \ref{fig-aligned_process}, each pair has one rendered image and one experimental image on the same depth. 
Of these, 70\% were allocated to the training set, 15\% to the validation set, and 15\% to the test set. The model is trained for 100 epochs.
The code was implemented in PyTorch 1.8.1 and Python 3.8, running on a system equipped with 1 NVIDIA A100 GPU with 80 GB of memory. The CUDA version used was 11.4, and the inference precision was set to float32.

\begin{table*}[t!]
\vspace*{1em} 
\centering
\caption{Performance comparison of different models. The best results in each column are depicted in boldface.}
\begin{tabular}{|c|c|c|c|c|}
\hline
Model/Value   & Time (s) ($\uparrow$) & SSIM ($\uparrow$)   & PSNR ($\uparrow$)   & MSE (e-02) ($\downarrow$)    \\ \hline\hline
GAN              & \textbf{0.002}   & 0.534   & 15.211  & 3.025 \\ \hline
Rendering        & 0.020   & 0.639   & 14.728  & 3.587 \\ \hline
Rendering + GAN  & 0.022   & \textbf{0.724}   & \textbf{18.370}  & \textbf{1.548} \\ \hline
\end{tabular}
\label{table:comparison}
\end{table*}

\begin{table*}[t!]
\centering
\caption{Pose estimation results using models trained on experimental (\textit{Exp}) and generated (\textit{Gen}) images.}
\label{table:pose}
\begin{tabular}{|c|c||c|c|c|c|c|c|c|c|}
\hline 
\multirow{2}{*}{Model}   &\multirow{2}{*}{Dataset} & \multicolumn{2}{c|}{Accuracy ($\uparrow$)} & \multicolumn{2}{c|}{Precision ($\uparrow$)}   & \multicolumn{2}{c|}{Recall ($\uparrow$)}   & \multicolumn{2}{c|}{F1 Score ($\uparrow$)}\\ 
\cline{3-10}
& & Pitch & Roll& Pitch & Roll& Pitch & Roll& Pitch & Roll\\ 
\hline\hline
\multirow{2}{*}{CNN}  & Exp& 0.988& 0.971&0.992 &0.981 & 0.992&0.978 & 0.992&0.978\\ \cline{2-10}
 & Gen& \underline{0.939} & \underline{0.919} & \underline{0.955}&\underline{0.916}  & \underline{0.954} & \underline{0.916} & \underline{0.952} & \underline{0.914} \\ \hline
\multirow{2}{*}{ResNet18}  & Exp & \textbf{0.991} & \textbf{1.000} & \textbf{0.993} & \textbf{1.000} & \textbf{0.994} &\textbf{1.000} &\textbf{0.993} & \textbf{1.000}\\ \cline{2-10}
 & Gen&  0.859 & 0.812  &  0.892 & 0.856  & 0.829  & 0.769  &  0.846 & 0.770  \\ \hline
\multirow{2}{*}{ViT}  & Exp& \textbf{0.991}& 0.977&0.987 & 0.978&0.984 & 0.975&0.985  &0.976\\ \cline{2-10}
 & Gen& 0.832 &  0.812&0.844  &0.819  & 0.757 & 0.779 & 0.772 & 0.781 \\ \hline
\end{tabular}\\
\textbf{Bold}: best among Exp models; \underline{Underlined}: best among Gen models.
\label{table-pose}
\vspace{-0.2cm}
\end{table*}


\subsection{Evaluation Metrics}
To evaluate the quality of the generated images, the work assessed both the image generation time and several quantitative metrics, including SSIM, PSNR, and MSE. 
MSE quantifies the average squared difference between pixel values of the ground-truth image ("target image") and the corresponding GAN-generated image. 
Lower MSE values indicate smaller pixel-wise differences and better alignment.
PSNR, expressed in decibels (dB), measures the peak error between target and generated images. Higher PSNR values denote better image quality.
SSIM evaluates structural similarity by analyzing luminance, contrast, and structural components. SSIM values range from 0 to 1, with values closer to 1 reflecting higher similarity and better visual quality.
Inference of the GAN model was performed on a system equipped with an NVIDIA A100 GPU.

\begin{figure}[t!]
\captionsetup{font=footnotesize,labelsep=period}
\centering
\includegraphics[width=1.02\hsize]{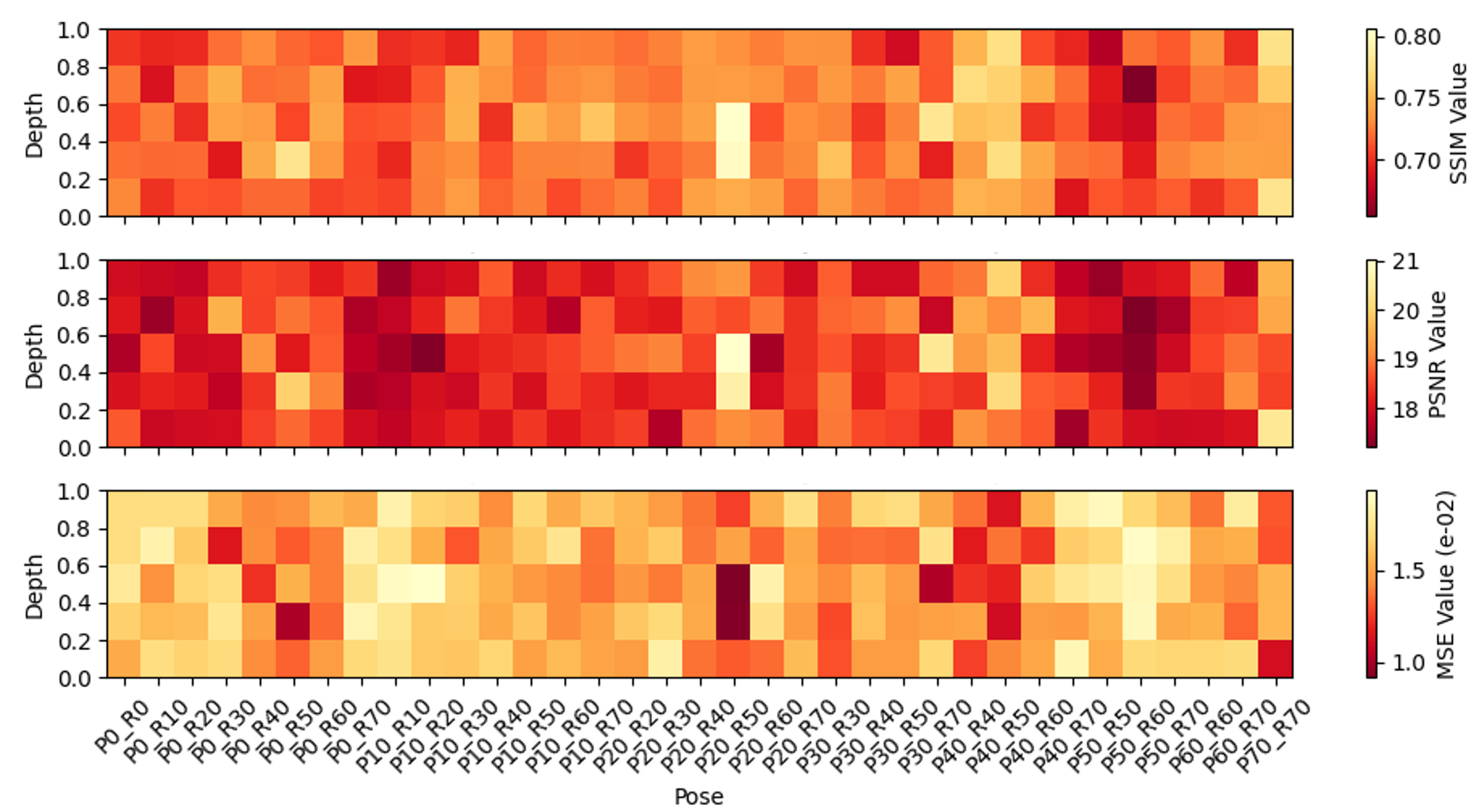}
\caption[Data Alignment]
{Heatmap of evaluation metrics (SSIM, PSNR, MSE) across different robot poses and depths. The X-axis represents the robot’s posture angles, while the Y-axis indicates the height offset relative to the focal plane. Each cell corresponds to a specific combination of pose and depth. The horizontal axis represents the robot’s posture angles, denoted as \texttt{P\textit{a}\_R\textit{b}}, where \textit{a} and \textit{b} indicate the pitch and roll angles in degrees, respectively (e.g., \texttt{P0\_R60} means pitch = 0$^\circ$ and roll = 60$^\circ$).}
\label{fig-result1}
 \vspace{-0.5cm}
\end{figure}%


\subsection{Image Generation Results}

Table~\ref{table:comparison} compares three approaches for generating the images:
1) a GAN that uses only the CAD image as input, denoted as `GAN';
2) a physics-based rendering method using the CAD image, denoted as `Rendering';
3) a hybrid approach where a GAN takes the physics-based rendering as input, denoted as `Rendering + GAN'.
These comparisons serve as ablation studies.

The results show that combining physics-based rendering with the GAN significantly improves the overall image quality (by 35\%) with only a minimal increase in per-image generation time (0.02 s).
Fig. \ref{fig-result} provides example images from the real experiment, as well as corresponding CAD, rendered, and GAN-generated outputs, while
Fig. \ref{fig-result1} presents heatmaps illustrating the performance across each of the 35 pose classes.
Notably, the GAN-generated images with physics-based rendering as input achieve the highest SSIM scores, surpassing those obtained from the physics-based rendering alone for each pose.

\subsection{Pose Estimation}
\subsubsection{Result} 
The work further evaluates the quality of the generated data in the downstream task of microrobot pose estimation (pitch and roll), with results presented in Table~\ref{table-pose}.
Four evaluation metrics are used: accuracy, precision, recall, and F1 score. 
To investigate how network depth and inductive bias influence microscopy-image pose estimation, three backbones are selected: convolutional-based architectures (CNN and ResNet18) for their strong spatial feature extraction capabilities, 
and transformer-based model (ViT~\cite{dosovitskiy2021imageworth16x16words}, which is good at capturing global dependencies.
These architectures span different architectures, enabling a controlled study of (a) network depth~(CNN vs ResNet-18) and (b) inductive bias~(convolutional locality vs. global self-attention) under our limited-scale dataset.

Each model is trained or fine-tuned for 30 epochs to predict pitch and roll angles.
The models are tested on the same experimental real data, consisting of 350 images representing 35 different poses. To avoid data leakage, which may inflate the pose estimation performance, the 350 test images were strictly excluded from the training sets of both the GAN and pose-estimation networks. 
Across three backbones, synthetic-data models underperform models trained on experimental data by 5.0\%--18.8\% in accuracy (pitch: 5.0\%--16.0\%; roll: 5.4\%--18.8\%).
For the strongest CNN, the accuracy gaps are modest (pitch/roll: 5.0\%/5.4\%), demonstrating the high similarity between the generated and real datasets; for precision/recall/F1 the relative drops are 3.7\%--4.0\% (pitch) vs 6.2\%--6.6\% (roll).

\subsubsection{Hybrid Training} 
To evaluate the quality of our generated images, we conducted experiments using hybrid datasets for pose estimation model training with CNN backbone. 
We constructed multiple training datasets by combining experiment images with generated images at different ratios: 100\% Exp, 75\% Exp + 25\% Gen, 50\% Exp + 50\% Gen, 25\% Exp + 75\% Gen, and 100\% Gen, while maintaining consistent test evaluation on 350 experiment images across all experiments. The test images were strictly excluded from the training sets of both the GAN and pose-estimation networks. Each configuration was tested three times, and the results were averaged to ensure statistical reliability.


The experimental results, presented in Table~\ref{table-hybrid}, demonstrate the effectiveness of our generated images in downstream pose estimation tasks. 

Overall in hybrid training, accuracy decreases as the generated-image ratio increases: relative to 100\% Exp, pitch drops by 0.7\%--5.0\% and roll by 2.3\%--5.4\%; precision/recall/F1 follow the same pattern (pitch \(\approx\) 0.7\%--4.0\%, roll \(\approx\)1.9\%--6.6\%).
Notably, replacing 50\% of experimental images reduces pitch accuracy by only 0.9\% versus 100\% Exp, and roll accuracy at 100\% Gen matches the 50\% Gen case (both 0.919), indicating that modest mixing preserves accuracy.

\begin{table*}[t!]
\centering
\caption{Pose estimation results using CNN trained on hybrid experimental (Exp) and generated (Gen) images.}
\begin{tabular}{|c||c|c|c|c|c|c|c|c|}
\hline 
\multirow{2}{*}{Hybrid Data} & \multicolumn{2}{c|}{Accuracy ($\uparrow$)} & \multicolumn{2}{c|}{Precision ($\uparrow$)}   & \multicolumn{2}{c|}{Recall ($\uparrow$)}   & \multicolumn{2}{c|}{F1 Score ($\uparrow$)}\\ 
\cline{2-9}
& Pitch & Roll& Pitch & Roll& Pitch & Roll& Pitch & Roll\\ 
\hline\hline
100\% Exp  &0.988& 0.971&0.992 &0.981 & 0.992&0.978 & 0.992&0.978\\ 
75\% Exp + 25\% Gen  & 0.981  & 0.949 & 0.985 &  0.962 &0.975  & 0.960 & 0.980 & 0.959 \\ 
50\% Exp + 50\% Gen  & 0.979  & 0.919 & 0.985 & 0.932  & 0.981 &0.924& 0.982 & 0.922 \\ 
25\% Exp + 75\% Gen  & 0.959  &0.921  & 0.966 &  0.919 & 0.955 & 0.949 & 0.965 & 0.951 \\ 
100\% Gen&  0.939 & 0.919 & 0.955 & 0.916  & 0.954 & 0.916 & 0.952 & 0.914
\\ 
\hline
\end{tabular}
\label{table-hybrid}
\end{table*}

\begin{table*}[t!]
\centering
\caption{Pose Estimation Performance on Generated Images from PixelGAN Models Trained with Different Pose Sets.
Average results over three experiments using the CNN-based pose estimation model. 
PixelGAN-35: trained on all 35 poses; PixelGAN-30: trained on Set B only (30 poses excluding Set A: P0\_R20, P10\_R30, P20\_R40, P30\_R50 and P40\_R60). 
Both models generated images for all 35 poses for the pose estimation model training.
}
\begin{tabular}{|c||c|c|c|c|c|c|c|c|}
\hline 
\multirow{2}{*}{Model} & \multicolumn{2}{c|}{Accuracy ($\uparrow$)} & \multicolumn{2}{c|}{Precision ($\uparrow$)}   & \multicolumn{2}{c|}{Recall ($\uparrow$)}   & \multicolumn{2}{c|}{F1 Score ($\uparrow$)}\\ 
\cline{2-9}
& Set A & Set B & Set A & Set B& Set A & Set B& Set A & Set B\\ 
\hline\hline
PixelGAN-35 &  0.888 & 0.938 & 0.855 & 0.945  & 0.820 & 0.933 & 0.828 & 0.937 \\ 
PixelGAN-30  & 0.866  & 0.916 & 0.707 & 0.915  & 0.661 & 0.896 & 0.675 & 0.899 \\  
\hline
\end{tabular}
\label{table-generalise}
\vspace{-0.2cm}
\end{table*}

\subsubsection{Generalisability}
To evaluate our model's capability in generating images for unseen poses, we conducted a generalisability experiment. 
We divided the 35 pose categories into Set~A (P0\_R20, P10\_R30, P20\_R40, P30\_R50 and P40\_R60) and Set~B (the remaining 30 poses). Set~A groups 5 challenging poses (hard to reach in physical experiments, structurally asymmetric, with strong optical artifacts), which makes them harder to acquire, to simulate faithfully, and to estimate accurately; this split provides a stringent generalisation test of PixelGAN on under-sampled, unseen cases.
We trained two PixelGAN models under different conditions: 1) PixelGAN-35: trained on the complete dataset (Set A + Set B), and 2) PixelGAN-30: trained exclusively on Set B data. 
Subsequently, both models were used to generate images for all 35 pose categories, which were then employed to train pose estimation models.

The pose estimation models trained on generated images were evaluated on their classification performance across different pose sets. 
The results, presented in Table~\ref{table-generalise}, demonstrate the generalisability. 
Set~A is consistently harder than Set~B across both generators, with accuracy lower by 5.3\%--5.5\%. Robustness to unseen poses is validated by comparing PixelGAN-30 with PixelGAN-35 on Set~A (unseen for PixelGAN-30): accuracy shows a 2.5\% relative drop (0.866 vs 0.888), which is smaller than the set’s inherent difficulty. The accuracy difference between the two models on seen poses (Set~B) is likewise small, at 2.4\% (0.916 vs 0.938). Other metrics follow the same trend: on Set~A, PixelGAN-30 vs PixelGAN-35 exhibits 0.148--0.159 absolute drops (precision/recall/F1), versus 3.2\%--4.1\% relative drops on Set~B.


\section{Discussion and Future Work}
This study presents a physics-informed deep generative learning framework integrating wave optics-based physical modeling {and depth alignment into} PixelGAN, significantly enhancing sim-to-real data augmentation for microrobot pose estimation. Unlike purely data-driven or physics-based methods, this hybrid strategy employs accurate physical rendering to guide PixelGAN in efficiently capturing fine-grained visual features, such as depth-encoded diffraction rings and subtle optical artifacts, with minimal computational overhead. Specifically, the pixel-wise correspondence in PixelGAN preserves structural coherence and mid-level visual details crucial for accurate pose and depth estimation.

Model performance varied by data source: pre-trained ResNet18 achieved optimal results on experimental images, while a simpler 3-layer CNN excelled on generated images. This suggests that synthetic images maintain essential pose features but exhibit more consistent, less complex distributions than real experimental data. Consequently, simpler CNN architectures suit the relatively uniform synthetic data better, reducing the risk of overfitting compared to more complex models. The ViT showed suboptimal performance across all datasets, which can be attributed to its need for larger datasets to unlock its full potential.

Although this approach substantially reduces the sim-to-real performance gap and significantly lowers the cost and complexity of experimental data collection, synthetic images still result in a minor performance difference (5.0\%--5.4\% with CNN pose estimator) compared to real-image-trained models. This residual gap arises partly from simplifications in the physics-based model and subtle variations in imaging conditions. Future research will address this by leveraging advanced domain adaptation techniques such as knowledge-guided transfer learning, feature-space alignment, and fine-tuning synthetic-trained models with limited experimental data. Further improvements to the physical simulation, including modeling complex optical phenomena like lens aberrations, sensor-specific noise, and fluidic diffraction effects, will enhance realism and robustness. Given the high rendering speed, exploiting this approach for RL-based microrobot control tasks could further enhance real-time interpretability, safety, and robustness in biomedical applications.

\section{Conclusion}
In summary, the work presents a physics-informed deep generative learning framework that significantly advances OM simulation for microrobot pose estimation. By combining wave-optics-based physical rendering with PixelGAN refinement, the method delivered a remarkable 35.6\% improvement in image fidelity (SSIM) and achieved rapid data generation (0.022 seconds/frame), substantially narrowing the sim-to-real gap. The high accuracy (within 5.0\%--5.4\% for the CNN pose estimator) confirms the viability of synthetic data for training robust microrobot vision algorithms, greatly reducing reliance on labour-intensive experimental datasets. Additionally, the framework demonstrates generalisability to unseen pose configurations, with an \(\sim\)2.5\% relative drop on unseen poses (accuracy), enabling robust deployment across diverse microrobotic scenarios.
This work thus offers a practical and interpretable solution for efficient dataset augmentation, laying the foundation for safer and more explainable biomedical microrobotic systems.

\bibliographystyle{IEEEtran}

\bibliography{references}

@article{kawata2001finer,
  title={Finer features for functional microdevices},
  author={Kawata, Satoshi and Sun, Hong-Bo and Tanaka, Tomokazu and Takada, Kenji},
  journal={Nature},
  volume={412},
  number={6848},
  pages={697--698},
  year={2001},
  publisher={Nature Publishing Group}
}

@article{4zhang2020distributed, title={Distributed force control for microrobot manipulation via planar multi-spot optical tweezer}, author={Zhang, Dandan and Barbot, Antoine and Lo, Benny and Yang, Guang-Zhong}, journal={Advanced Optical Materials}, volume={8}, number={21}, pages={2000543}, year={2020}, publisher={Wiley Online Library} }

@article{19rogers2012super,
  title={A super-oscillatory lens optical microscope for subwavelength imaging},
  author={Rogers, Edward TF and Lindberg, Jari and Roy, Tapashree and Savo, Salvatore and Chad, John E and Dennis, Mark R and Zheludev, Nikolay I},
  journal={Nature materials},
  volume={11},
  number={5},
  pages={432--435},
  year={2012},
  publisher={Nature Publishing Group UK London}
}

@article{20nasse2010realistic, title={Realistic modeling of the illumination point spread function in confocal scanning optical microscopy}, author={Nasse, Michael J and Woehl, J{\"o}rg C}, journal={Josa a}, volume={27}, number={2}, pages={295--302}, year={2010}, publisher={Optica Publishing Group} }

@article{21marian2007complex, title={On the complex three-dimensional amplitude point spread function of lenses and microscope objectives: theoretical aspects, simulations and measurements by digital holography}, author={Marian, Anca and Charri{\`e}re, Florian and Colomb, Tristan and Montfort, Fr{\'e}d{\'e}ric and K{\"u}hn, Jonas and Marquet, Pierre and Depeursinge, Christian}, journal={Journal of microscopy}, volume={225}, number={2}, pages={156--169}, year={2007}, publisher={Wiley Online Library} }

@article{22wang2011optical,
  title={Optical virtual imaging at 50 nm lateral resolution with a white-light nanoscope},
  author={Wang, Zengbo and Guo, Wei and Li, Lin and Luk'Yanchuk, Boris and Khan, Ashfaq and Liu, Zhu and Chen, Zaichun and Hong, Minghui},
  journal={Nature communications},
  volume={2},
  number={1},
  pages={218},
  year={2011},
  publisher={Nature Publishing Group UK London}
}

@article{23zhang2023large,
  title={Large depth-of-field ultra-compact microscope by progressive optimization and deep learning},
  author={Zhang, Yuanlong and Song, Xiaofei and Xie, Jiachen and Hu, Jing and Chen, Jiawei and Li, Xiang and Zhang, Haiyu and Zhou, Qiqun and Yuan, Lekang and Kong, Chui and others},
  journal={Nature Communications},
  volume={14},
  number={1},
  pages={4118},
  year={2023},
  publisher={Nature Publishing Group UK London}
}

@article{24li2024statistical,
  title={A statistical resolution measure of fluorescence microscopy with finite photons},
  author={Li, Yilun and Huang, Fang},
  journal={Nature Communications},
  volume={15},
  number={1},
  pages={3760},
  year={2024},
  publisher={Nature Publishing Group UK London}
}

@inproceedings{isola2017image,
  title={Image-to-image translation with conditional adversarial networks},
  author={Isola, Phillip and Zhu, Jun-Yan and Zhou, Tinghui and Efros, Alexei A},
  booktitle={Proceedings of the IEEE conference on computer vision and pattern recognition},
  pages={1125--1134},
  year={2017}
}

@article{wei2025dataset,
  title={A Dataset and Benchmarks for Deep Learning-Based Optical Microrobot Pose and Depth Perception},
  author={Wei, Lan and Zhang, Dandan},
  journal={arXiv preprint arXiv:2505.18303},
  year={2025}
}

@article{dosovitskiy2021imageworth16x16words,
  title={An image is worth 16x16 words: Transformers for image recognition at scale},
  author={Dosovitskiy, Alexey and Beyer, Lucas and Kolesnikov, Alexander and Weissenborn, Dirk and Zhai, Xiaohua and Unterthiner, Thomas and Dehghani, Mostafa and Minderer, Matthias and Heigold, Georg and Gelly, Sylvain and others},
  journal={arXiv preprint arXiv:2010.11929},
  year={2020}
}

@article{zhang2020data,
  title={Data-driven microscopic pose and depth estimation for optical microrobot manipulation},
  author={Zhang, Dandan and Lo, Frank P-W and Zheng, Jian-Qing and Bai, Wenjia and Yang, Guang-Zhong and Lo, Benny},
  journal={Acs Photonics},
  volume={7},
  number={11},
  pages={3003--3014},
  year={2020},
  publisher={ACS Publications}
}

@article{yang2024machine,
  title={Machine learning for micro-and nanorobots},
  author={Yang, Lidong and Jiang, Jialin and Ji, Fengtong and Li, Yangmin and Yung, Kai-Leung and Ferreira, Antoine and Zhang, Li},
  journal={Nature Machine Intelligence},
  volume={6},
  number={6},
  pages={605--618},
  year={2024},
  publisher={Nature Publishing Group UK London}
}

@article{zhang2022micro,
  title={Micro-object pose estimation with sim-to-real transfer learning using small dataset},
  author={Zhang, Dandan and Barbot, Antoine and Seichepine, Florent and Lo, Frank P-W and Bai, Wenjia and Yang, Guang-Zhong and Lo, Benny},
  journal={Communications Physics},
  volume={5},
  number={1},
  pages={80},
  year={2022},
  publisher={Nature Publishing Group UK London}
}

@article{zhang2022fabrication,
  title={Fabrication and optical manipulation of micro-robots for biomedical applications},
  author={Zhang, Dandan and Ren, Yunxiao and Barbot, Antoine and Seichepine, Florent and Lo, Benny and Ma, Zhuo-Chen and Yang, Guang-Zhong},
  journal={Matter},
  volume={5},
  number={10},
  pages={3135--3160},
  year={2022},
  publisher={Elsevier}
}

@article{wang2007laplacian,
  title={Laplacian operator-based edge detectors},
  author={Wang, Xin},
  journal={IEEE transactions on pattern analysis and machine intelligence},
  volume={29},
  number={5},
  pages={886--890},
  year={2007},
  publisher={IEEE}
}

@article{grier2003revolution,
  title={A revolution in optical manipulation},
  author={Grier, David G},
  journal={nature},
  volume={424},
  number={6950},
  pages={810--816},
  year={2003},
  publisher={Nature Publishing Group UK London}
}

@article{sha2019review,
  title={A review on microscopic visual servoing for micromanipulation systems: Applications in micromanufacturing, biological injection, and nanosensor assembly},
  author={Sha, Xiaopeng and Sun, Hui and Zhao, Yuliang and Li, Wenchao and Li, Wen J},
  journal={Micromachines},
  volume={10},
  number={12},
  pages={843},
  year={2019},
  publisher={MDPI}
}

@article{muinos2021reinforcement,
  title={Reinforcement learning with artificial microswimmers},
  author={Mui{\~n}os-Landin, Santiago and Fischer, Alexander and Holubec, Viktor and Cichos, Frank},
  journal={Science Robotics},
  volume={6},
  number={52},
  pages={eabd9285},
  year={2021},
  publisher={American Association for the Advancement of Science}
}

@article{shurrab2022self,
  title={Self-supervised learning methods and applications in medical imaging analysis: A survey},
  author={Shurrab, Saeed and Duwairi, Rehab},
  journal={PeerJ Computer Science},
  volume={8},
  pages={e1045},
  year={2022},
  publisher={PeerJ Inc.}
}

@article{plompen2020joint,
  title={The joint evaluated fission and fusion nuclear data library, JEFF-3.3},
  author={Plompen, Arjan JM and Cabellos, O and de Saint Jean, Cyrille and Fleming, Michael and Algora, A and Angelone, Maurizio and Archier, P and Bauge, E and Bersillon, O and Blokhin, A and others},
  journal={The European Physical Journal A},
  volume={56},
  pages={1--108},
  year={2020},
  publisher={Springer}
}

@article{li2024control,
  title={Control of self-winding microrobot using an electromagnetic drive system: integration of movable electromagnetic coil and permanent magnet},
  author={Li, Hao and Zhang, Zhaopeng and Yi, Xin and Jin, Shanhai and Chen, Yuan},
  journal={Micromachines},
  volume={15},
  number={4},
  pages={438},
  year={2024},
  publisher={MDPI}
}

@article{balakrishnan2023single,
  title={Single-shot, coherent, pop-out 3D metrology},
  author={Balakrishnan, Deepan and Chee, See Wee and Baraissov, Zhaslan and Bosman, Michel and Mirsaidov, Utkur and Loh, N Duane},
  journal={Communications Physics},
  volume={6},
  number={1},
  pages={321},
  year={2023},
  publisher={Nature Publishing Group UK London}
}

@article{grammatikopoulou2019three,
  title={Three-dimensional pose estimation of optically transparent microrobots},
  author={Grammatikopoulou, Maria and Yang, Guang-Zhong},
  journal={IEEE Robotics and Automation Letters},
  volume={5},
  number={1},
  pages={72--79},
  year={2019},
  publisher={IEEE}
}

@inproceedings{cenev2016object,
  title={Object tracking in robotic micromanipulation by supervised ensemble learning classifier},
  author={Cenev, Zoran and Ven{\"a}l{\"a}inen, Janne and Sariola, Veikko and Zhou, Quan},
  booktitle={2016 International Conference on Manipulation, Automation and Robotics at Small Scales (MARSS)},
  pages={1--5},
  year={2016},
  organization={IEEE}
}

@inproceedings{ren2022machine,
  title={Machine learning-based real-time localization and automatic trapping of multiple microrobots in optical tweezer},
  author={Ren, Yunxiao and Keshavarz, Meysam and Anastasova, Salzitsa and Hatami, Ghazal and Lo, Benny and Zhang, Dandan},
  booktitle={2022 international conference on manipulation, automation and robotics at small scales (MARSS)},
  pages={1--6},
  year={2022},
  organization={IEEE}
}

@article{dong2024ai,
  title={AI-enhanced biomedical micro/nanorobots in microfluidics},
  author={Dong, Hui and Lin, Jiawen and Tao, Yihui and Jia, Yuan and Sun, Lining and Li, Wen Jung and Sun, Hao},
  journal={Lab on a Chip},
  volume={24},
  number={5},
  pages={1419--1440},
  year={2024},
  publisher={Royal Society of Chemistry}
}

@article{wei2025physics,
  title={Physics-Informed Machine Learning with Adaptive Grids for Optical Microrobot Depth Estimation},
  author={Wei, Lan and Genoud, Lou and Zhang, Dandan},
  journal={arXiv preprint arXiv:2509.02343},
  year={2025}
}
\end{document}